\documentclass[conference,10pt]{IEEEtran}
\usepackage{graphicx}
\usepackage{epstopdf}
\usepackage{graphicx}
\usepackage{amsmath}
\usepackage{amsfonts}
\usepackage{amssymb, color, bm}

\usepackage{algorithmicx}
\usepackage[noend]{algpseudocode}
\usepackage{algorithm,setspace}
\usepackage[T1]{fontenc}

\usepackage{multicol}
\usepackage{lipsum}
\usepackage{mwe}

\DeclareMathOperator{\E}{\mathbb{E}}

\usepackage[utf8]{inputenc}
\allowdisplaybreaks
\usepackage{cite}

\usepackage{algorithm}
\usepackage[noend]{algpseudocode}

\usepackage{lineno}

\title{Bayesian Active Meta-Learning for Black-Box Optimization}
\author{
    \IEEEauthorblockN{Ivana Nikoloska and Osvaldo Simeone}
    \IEEEauthorblockA{KCLIP, CTR, Dept. of Engineering, King's College London
    \\\{ivana.nikoloska, osvaldo.simeone\}@kcl.ac.uk}
    }

\begin{document}

\maketitle
\let\thefootnote\relax\footnote{This work was supported by the European Research Council (ERC) under the European Union’s Horizon 2020 Research and Innovation Program (Grant
Agreement No. 725731).}

\begin{abstract}
Data-efficient learning algorithms are essential in many practical applications for which data collection is expensive, e.g., for the optimal deployment of wireless systems in unknown propagation scenarios. Meta-learning can address this problem by leveraging data from a set of related learning tasks, e.g., from similar deployment settings. In practice, one may have available only unlabeled data sets from the related tasks, requiring a costly labeling procedure to be carried out before use in meta-learning. For instance, one may know the possible positions of base stations in a given area, but not the performance indicators achievable with each deployment. To decrease the number of labeling steps required for meta-learning, this paper introduces an information-theoretic active task selection mechanism, and evaluates an instantiation of the approach for Bayesian optimization of black-box models. 
\end{abstract}


\section{Introduction}
One of the key applications of artificial intelligence (AI) to engineering concerns is the optimization of black-box functions representing the end-to-end performance of a complex system. Examples include optimal resource allocation for the deployment of wireless systems \cite{maggi2021bayesian} and the optimization of electronic circuits for power amplifiers \cite{yousefzadeh2006band}. In these settings, the objective function depends on the interaction of many components, and is often analytically intractable and expensive to evaluate, requiring separate deployments or experiments to probe the function at different input values. 

\emph{Bayesian Optimization (BO)} provides a principled technique to address global black-box optimization  \cite{frazier2018bayesian}, \cite{shahriari2015taking}. BO is a supervised learning method that treats the objective function as an unknown function, sampled from a \emph{Gaussian process (GP)}, that is observed through noisy measurements. BO is often coupled with \emph{active learning}, whereby the next input at which to query the objective function is selected using an acquisition function that depends on the data collected so far and on the resulting uncertainty associated with different inputs.

In BO, the GP serves as a prior on the objective function, and is defined by a kernel function that specifies the expected level of smoothness of the function. The kernel is typically selected a priori based on domain knowledge about the problem under study. A prior that is better suited to model the actual objective function yields lower requirements on the amount of data needed to obtain given performance levels \cite{rasmussen2003gaussian}.



While the prior cannot be optimized based on data from the task of interest, it is possible to use data from \emph{related learning tasks}, if available, to make a more informed choice about the prior.  Optimizing the prior based on data from related tasks is the goal of \emph{meta-learning}, which amounts to empirical Bayes in this context \cite{grant2018recasting}.   Meta-learning operates by extracting shared knowledge, in the form of a shared prior, from data collected from multiple tasks \cite{grant2018recasting,rothfuss2021pacoh}. 

In the standard formulation of meta-learning, the data sets corresponding to the related tasks are typically assumed to be labeled and ready for use. In practice however, one may have available only unlabeled data sets from the tasks, requiring a costly labeling procedure to be carried out, before use in standard meta-learning schemes.

\begin{figure*}[tbp]
\centering
\includegraphics[width=0.52\linewidth]{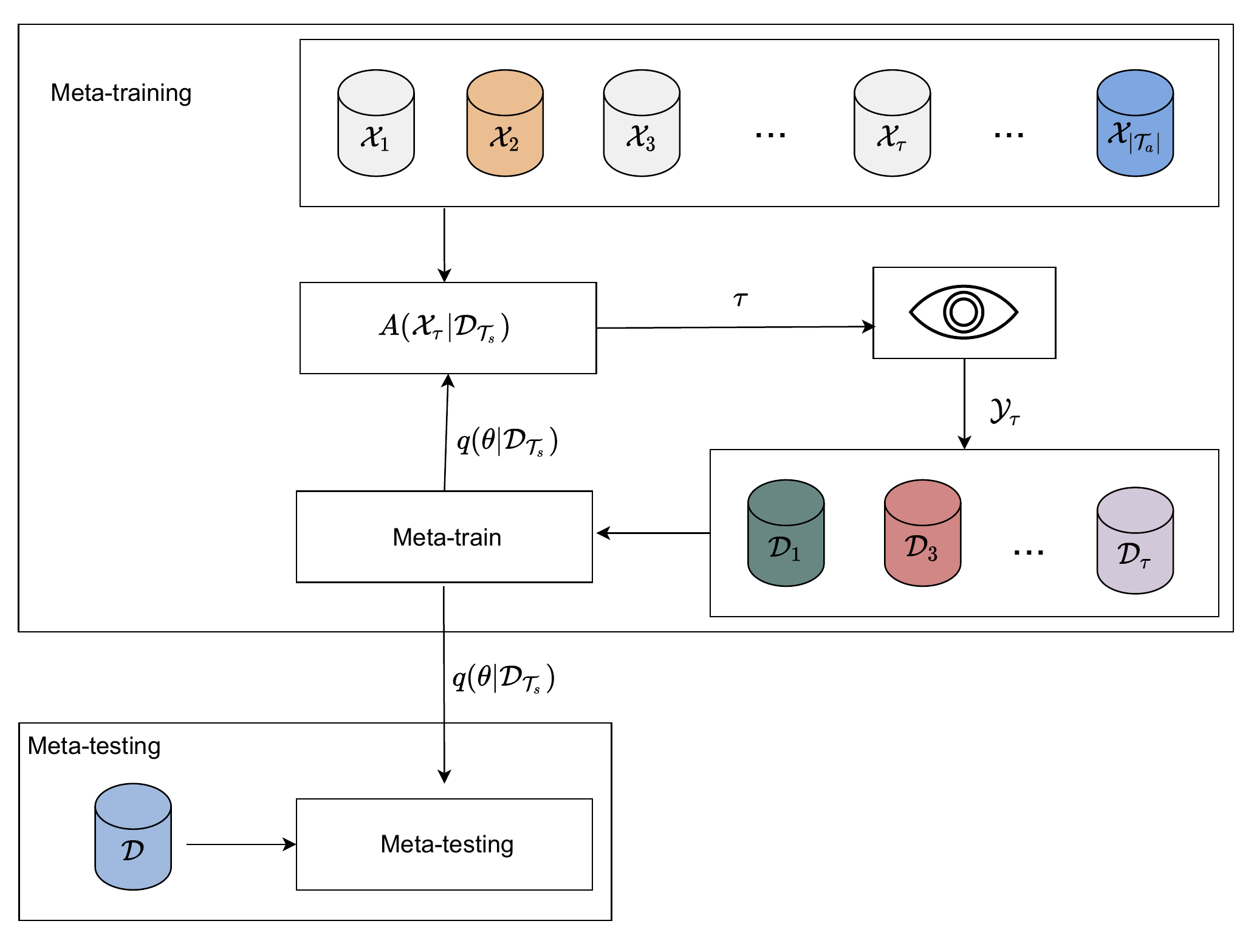}
\caption{Illustration of the active meta-learning setting under study. The meta-learner optimizes the variational posterior $q(\theta|\mathcal{D}_{\mathcal{T}_s})$ of the shared hyperparameters $\theta$ by using data from the set $\mathcal{T}_s$ of selected meta-training tasks in order to enable fast adaptation to a new task during meta-testing. The problem of interest is how to select sequentially the next task $\tau$ to add to the set $\mathcal{T}_s$ of selected tasks from a pool $\mathcal{T}_a \setminus \mathcal{T}_s$ of meta-training tasks for which we only have accessed to unlabeled data. Upon selection of task $\tau$, the labels $\mathcal{Y}_{\tau}$ are obtained by the meta-learner for the given available covariate vectors $\mathcal{X}_{\tau}$.}
\label{sys}
\end{figure*}

As shown in Fig.~\ref{sys}, this paper addresses a problem formulation motivated by settings in which the learning agent has unlabeled data sets for multiple related tasks. We focus on devising strategies for the active selection of which tasks should be selected for labeling, say by a human annotator, so as to enhance the BO efficacy. The problem is formulated as \emph{active meta-learning} within a hierarchical Bayesian framework in which a vector of hyperparameters is shared among the learning tasks. The proposed approach attempts to sequentially maximize the amount of information that the learner has about the hyperparameters by actively selecting which task should be labeled next.

The methodology developed in this paper extends the principles introduced in the vast literature on  \textit{active learning} \cite{felder2009active} to the higher-level problem of meta-learning, with a specific focus on black-box optimization. Active learning refers to the problem of selecting \emph{examples} to label so as to improve the sample efficiency of a training algorithm in inferring \emph{model parameters}. In contrast, active meta-learning selects \emph{data sets} corresponding to distinct \emph{learning tasks} in order to infer more efficiently \emph{hyperpameters} with the goal of generalizing more quickly to new tasks.

\subsection{Notation} 
We will use $\mathrm{H}(\cdot \, | \, \cdot)$ to denote the conditional differential entropy, which for simplicity will be referred to as entropy; and we use $\mathrm{I}(\cdot;\cdot \, | \, \cdot)$ to denote the conditional mutual information.

\section{Information-Theoretic Active Meta-Learning}

\subsection{Setting}
In the active meta-learning setting under study, shown in Fig.~\ref{sys}, the meta-learner aims at optimizing a \emph{hyperparameter} vector $\theta$ as to maximize the performance of the latter on a new, a priori unknown, task by using data from a small number of related meta-training tasks. In the setting of interest, the hyperparameter $\theta$ defines the kernel of a GP, but we will keep the presentation more general here. At the beginning of the meta-learning process, the meta-learner has access to \emph{unlabeled} data sets ${\mathcal{D}}_{\mathcal{T}_a} = \{ ({\mathcal{X}}_\tau)\}_{\tau = 1}^{|\mathcal{T}_a|}$ from a set $\mathcal{T}_a = \{1,...,|\mathcal{T}_a|\}$ of $|\mathcal{T}_a|$ \emph{meta-training tasks}. Each data set ${\mathcal{X}}_\tau = \{({x}_\tau^n)\}_{n=1}^{N_\tau}$ comprises $N_\tau$ covariate vectors, with no labels. At each step of the meta-learning process, the meta-learner chooses the next task to be labeled in a sequential fashion.  When the meta-learner selects a task $\tau$, it receives the corresponding labels ${\mathcal{Y}}_\tau = \{({y}_\tau^n)\}_{n=1}^{N_\tau}$, obtaining the complete data set $\mathcal{D}_{\tau}=({\mathcal{X}}_\tau,{\mathcal{Y}}_\tau)$. 

At each step of the selection process, the meta-learner has obtained labels for a subset $\mathcal{T}_s \subseteq \mathcal{T}_a$ of meta-training tasks, and hence it has access to the meta-training data set ${\mathcal{D}}_{\mathcal{T}_s} = \{ ({\mathcal{X}}_\tau,{\mathcal{Y}}_\tau)\}_{\tau = 1}^{|\mathcal{T}_s|}$. The next task to be labeled in the set $\mathcal{T}_a \setminus \mathcal{T}_s$ is selected based on a \emph{meta-acquisition function} $\mathrm{A}({\mathcal{X}}_\tau \, | \, {\mathcal{D}}_{\mathcal{T}_s})$ that assigns a score to any of the remaining meta-training tasks in the set $\mathcal{T}_a \setminus \mathcal{T}_s$ as
\begin{align}\label{acqui}
    \underset{\tau \in \mathcal{T}_a \setminus \mathcal{T}_s}{\text{arg max}} \left\{  \mathrm{A}(\tilde{\mathcal{X}}_\tau \, | \, {\mathcal{D}}_{\mathcal{T}_s}) \right\}.
\end{align}
In \eqref{acqui}, we have allowed for the use of a subset  $\tilde{{\mathcal{X}}}_\tau \subseteq {{\mathcal{X}}}_\tau$ of the available data for the meta-training task $\tau$ in order to control complexity. Upon selection of a candidate task $\tau$, the corresponding labeled data set $\mathcal{D}_{\tau}$ is incorporated in the meta-training data set, and the procedure is repeated until the budget $B\leq |\mathcal{T}_a|$ of meta-training tasks is exhausted. 

We assume a discriminative model defined by a conditional distribution $p(y|x, \phi)$ of output $y$ given the input, or covariates, $x$ and the model parameters $\phi$. In the setting of interest, the input $x$ corresponds to a vector of variables under optimization, e.g., the configuration of a wireless system, and $y$ represents the noisy observation of the objective function. We describe the model using a parametric density $p(y|x, \phi)$, but the derivation applies directly to non-parametric models such as GPs (see, e.g., \cite{houlsby2011bayesian}).

We adopt a hierarchical Bayesian model whereby the model parameters for each task ${\phi}_\tau$ have a shared prior distribution $p({\phi}_\tau\,|\,{\theta})$ determined by the common hyperparameter $\theta$. This model captures the statistical relationship between the meta-training tasks, and provides a useful way to derive principled meta-learning algorithms \cite{rothfuss2021pacoh}, \cite{jose2021epistemic}. Both the model parameters
${\phi}_\tau$ and hyperparameters ${\theta}$ are assumed to be latent random variables, with the joint distribution factorizing as $p(\{{\phi}_\tau\}_{\tau \in \mathcal{T}} , {\theta}) = p({\theta}) \prod_{\tau \in \mathcal{T}} p({\phi}_\tau \, | \, {\theta})$, with $p({\theta})$ denoting the prior distribution of the hyperparameters and $\mathcal{T}$ denoting any set of tasks. 

Combining prior and likelihood, the joint distribution of outputs  $\{\mathcal{Y}_\tau\}_{\tau \in \mathcal{T}}$, per-task model parameters $\{\phi_\tau\}_{\tau \in \mathcal{T}}$, and hyperparameter $\theta$, when conditioned on the corresponding covariates ${\mathcal{X}}_\tau$, takes the form
\begin{align}\label{joint}
    &p(\{\mathcal{Y}_\tau\}_{\tau \in \mathcal{T}}, \{\phi_\tau\}_{\tau \in \mathcal{T}}, \theta \, | \, {\mathcal{X}}_\tau) 
    \nonumber\\
    &= p(\theta)  \prod_{\tau \in \mathcal{T}} \left( p({\phi}_\tau \, | \, \theta) \prod_{n=1}^{N_\tau} p(y_\tau^n \, | \, {x}^n_\tau, {\phi}_\tau)\right). 
\end{align}
From \eqref{joint}, when the outputs for the meta-training tasks in set $\mathcal{T}_s$ have been acquired, the posterior distribution of any subset of the labels for any task $\tau \in \mathcal{T}_a \setminus \mathcal{T}_s$ is given by
\begin{align}\label{posterior_y}
    p (\tilde{\mathcal{Y}}_\tau \,|\, \tilde{{\mathcal{X}}}_\tau, {\mathcal{D}}_{\mathcal{T}_s}) = \E_{p({\theta} \,|\, {\mathcal{D}}_{\mathcal{T}_s})p({\phi}_\tau \,|\, {\theta})} [p(\tilde{\mathcal{Y}}_\tau \,|\, \tilde{{\mathcal{X}}}_\tau, {\phi}_\tau)],
\end{align}  where the average is taken over the marginal of the model parameter $\phi_{\tau}$ with respect to the joint posterior distribution $p({\theta},{\phi}_\tau \,|\, {\mathcal{D}}_{\mathcal{T}_s}) = p({\theta} \,|\, {\mathcal{D}}_{\mathcal{T}_s})p({\phi}_\tau \,|\, {\theta})$.

\subsection{Epistemic Uncertainty and Bayesian Active Meta-learning by Disagreement}
Given the current available data set ${\mathcal{D}}_{\mathcal{T}_s}$, under the assumed model (\ref{joint}), the overall average predictive uncertainty for a batch $\tilde{\mathcal{X}}_\tau$ of covariates for a candidate meta-training task $\tau \in \mathcal{T}_a \setminus \mathcal{T}_s$ can be measured by the conditional entropy \cite{houlsby2011bayesian}, \cite{xu2020minimum}, \cite{jose2021epistemic}
\begin{align}\label{pred_entropy}
    \mathrm{H}(\tilde{\mathcal{Y}}_\tau \,|\, \tilde{\mathcal{X}}_\tau,  {\mathcal{D}}_{\mathcal{T}_s}) = \E_{p(\tilde{\mathcal{Y}}_\tau \,|\, \tilde{\mathcal{X}}_\tau,  {\mathcal{D}}_{\tau_s})} [-\log p(\tilde{\mathcal{Y}}_\tau \,|\, \tilde{\mathcal{X}}_\tau,  {\mathcal{D}}_{\mathcal{T}_s})] ,
\end{align}
where the predictive posterior $p(\tilde{\mathcal{Y}}_\tau \,|\, \tilde{\mathcal{X}}_\tau,  {\mathcal{D}}_{\mathcal{T}_s})$ is given by \eqref{posterior_y}. The predictive uncertainty \eqref{pred_entropy} can be decomposed as \cite{jose2021epistemic}
\begin{align}\label{memr}
    \mathrm{H}(\tilde{\mathcal{Y}}_\tau \,|\, \tilde{\mathcal{X}}_\tau, {\mathcal{D}}_{\mathcal{T}_s}) = &\mathrm{H}  (\tilde{\mathcal{Y}}_\tau \,|\, \tilde{\mathcal{X}}_\tau, {\theta}, {\mathcal{D}}_{\mathcal{T}_s}) \nonumber\\
    &+ \mathrm{I} (\tilde{\mathcal{Y}}_\tau; {\theta} \,|\, \tilde{\mathcal{X}}_\tau, {\mathcal{D}}_{\mathcal{T}_s}).
\end{align}
As we will discuss next, the first term in (\ref{memr}) is a measure of \emph{aleatoric uncertainty}, which does not depend on the amount of labeled meta-training data; while the second is a measure of \emph{epistemic uncertainty}, which captures the informativeness of the labels for the new task $\tau$ given the already labeled data sets $\mathcal{D}_{\mathcal{T}_s}$. Based on this observation, we will propose to adopt (an estimate of) the latter term as a meta-acquisition function to be used in (\ref{acqui}).

To elaborate, let us first consider the entropy term in (\ref{memr}), which can be written as
\begin{align}\label{aleat}
    &\mathrm{H}(\tilde{\mathcal{Y}}_\tau \,|\, \tilde{\mathcal{X}}_\tau, {\theta}, {\mathcal{D}}_{\mathcal{T}_s}) =
    \E_{p({\theta} \,|\, {\mathcal{D}}_{\mathcal{T}_s})} [\mathrm{H}(\tilde{\mathcal{Y}}_\tau \,|\, \tilde{{\mathcal{X}}}_\tau, \theta)] \nonumber\\
    &= \E_{p({\theta} \,|\, {\mathcal{D}}_{\tau_s})p(\tilde{\mathcal{Y}}_\tau \,|\, \tilde{\mathcal{X}}_\tau, {\theta})} [-\log p(\tilde{\mathcal{Y}}_\tau \,|\, \tilde{\mathcal{X}}_\tau, {\theta})],
\end{align}
in which $p(\tilde{\mathcal{Y}}_\tau \,|\, {\tilde{\mathcal{X}}}_\tau, {\theta}) = \E_{p({\phi}_\tau \,|\, {\theta})} [p(\tilde{\mathcal{Y}}_\tau \,|\, \tilde{{\mathcal{X}}}_\tau, {\phi}_\tau)]$ is the predictive distribution under the assumed model (\ref{joint}) in the reference case in which one had access to the true hyperparameter vector $\theta$. Accordingly, the first term in \eqref{memr} captures the aleatoric uncertainty in the prediction on a new, a priori unknown, task $\tau$ that stems from the inherent randomness in the task generation. This term can not be reduced by collecting data from meta-training tasks, since such data can only improve the estimate of the hyperparameter vector $\theta$,  and is thus irrelevant for the purposes of task selection. We note that, the conditional entropy term in \eqref{aleat} is distinct from the standard aleatoric uncertainty in conventional learning in which the model parameter $\phi_\tau$, and not the hyperparameter $\theta$, is assumed to be unknown \cite{houlsby2011bayesian}.  

The second term in \eqref{memr} can be written as 
\begin{align}\label{MI_bamld_gen}
    &\mathrm{I}(\tilde{\mathcal{Y}}_\tau; {\theta} \,|\, \tilde{{\mathcal{X}}}_\tau, {\mathcal{D}}_{\mathcal{T}_s}) \nonumber\\
    &= \mathrm{H} (\tilde{\mathcal{Y}}_\tau \,|\, \tilde{\mathcal{X}}_\tau, \mathcal{D}_{\mathcal{T}_s}) - \mathrm{H}  (\tilde{\mathcal{Y}}_\tau \,|\, \tilde{\mathcal{X}}_\tau, \theta, \mathcal{D}_{\mathcal{T}_s})\nonumber\\
    &= \mathrm{H} (\tilde{\mathcal{Y}}_\tau \,|\, \tilde{{\mathcal{X}}}_\tau, {\mathcal{D}}_{\mathcal{T}_s}) - \E_{p({\theta} \,|\, {\mathcal{D}}_{\mathcal{T}_s})} [\mathrm{H}  (\tilde{\mathcal{Y}}_\tau \,|\, \tilde{{\mathcal{X}}}_\tau, {\theta})].
\end{align}
The mutual information (\ref{MI_bamld_gen}) is hence the difference between two terms: the overall predictive uncertainty (\ref{pred_entropy}) and the predictive (aleatoric) uncertainty (\ref{aleat}) corresponding to the ideal case in which the hyperparameter vector $\theta$ is known.  Intuitively, the conditional mutual information in \eqref{memr} represents the epistemic uncertainty arising from the limitations in the amount of available meta-training data, which causes the labels $\tilde{\mathcal{Y}}_\tau$ of a new meta-training task $\tau \in \mathcal{T}_a \setminus \mathcal{T}_s$ to carry useful information about the hyperparameter vector $\theta$ given the knowledge of the meta-training data set ${\mathcal{D}}_{\mathcal{T}_s}$.


Given that the aleatoric uncertainty, the first term in (\ref{memr}), is independent of the amount of the selected meta-training data, we propose that the meta-learner select tasks that minimize the epistemic uncertainty, i.e., the second term, in \eqref{memr}. 
The corresponding meta-acquisition problem (\ref{acqui}) is given as
\begin{align}\label{bamld_gen}
    \underset{\tau \in \mathcal{T}_a \setminus \mathcal{T}_s}{\text{arg max}} \left\{ \mathrm{I} (\tilde{\mathcal{Y}}_\tau; {\theta} \,|\, \tilde{{\mathcal{X}}}_\tau, {\mathcal{D}}_{\mathcal{T}_s})\right\}.
\end{align} The resulting Bayesian Active Meta-learning by Disagreement (BAMLD) algorithm is summarized in Algorithm 1.
 

To further interpret the proposed meta-acquisition function, let us consider again the decomposition (\ref{MI_bamld_gen}) of the conditional mutual information. The first term in \eqref{MI_bamld_gen} corresponds to the average log-loss of a predictor that averages over the posterior distribution $p({\theta} \,|\, {\mathcal{D}}_{\mathcal{T}_s})$ of the hyperparameter, whilst the second corresponds to the average log-loss of a genie-aided prediction that knows the correct realisation of the hyperparameter $\theta$, assuming a well-specified model. These two terms differ more significantly, making the conditional mutual information larger, when distinct choices of the hyperparameter vector $\theta$ yield markedly different predictive distributions $p(\tilde{\mathcal{Y}}_\tau \,|\, \tilde{{\mathcal{X}}}_\tau, {\theta})$. 
Thereby, the meta-learner does not merely select tasks with maximal predictive uncertainty, but rather it chooses tasks $\tau$ for which individual choices of the hyperparameter vector $\theta$ ``disagree'' more significantly.

BAMLD is the natural counterpart of the Bayesian Active Learning by Disagreement (BALD) method, introduced in \cite{houlsby2011bayesian} for conventional learning. Crucially, while BALD gauges disagreement at the level of model parameters, BAMLD operates at the level of hyperparameters, marginalizing out the model parameters.

\subsection{Implementation of BAMLD via Variational Inference}
The posterior distribution $p(\theta|\mathcal{D}_{\mathcal{T}_s})$ of the hyperparameters given the selected data $\mathcal{D}_{\mathcal{T}_s}$, which is needed to evaluate the two terms in BAMLD meta-acquisition function \eqref{bamld_gen} (see \eqref{MI_bamld_gen}), is generally intractable. To address this problem, we assume that an approximation of the posterior has been obtained via standard variational inference (VI) methods, yielding a variational distribution $q(\theta \,|\, \mathcal{D}_{\mathcal{T}_s})$ (see, e.g., \cite{angelino2016patterns}). Using the variational distribution, the first and second term in \eqref{MI_bamld_gen} can be estimated by replacing the true posterior $p(\theta|\mathcal{D}_{\mathcal{T}_s})$ with the variational distribution $q(\theta \,|\, \mathcal{D}_{\mathcal{T}_s})$. In practice, averages over the variational distribution $q(\theta \,|\, \mathcal{D}_{\tau_s})$ can be further estimated using Monte Carlo sampling. Alternatively, samples approximately distributed according to the true posterior distribution $p(\theta|\mathcal{D}_{\mathcal{T}_s})$ can be obtained via Monte Carlo methods, such as Stochastic-Gradient Markov Chain Monte Carlo (see, e.g., \cite{angelino2016patterns}), or via particle-based methods like Stein Variational Gradient Descent (SVGD) \cite{liu2016stein}.

\begin{algorithm}
\caption{Active meta-learning via BAMLD}

\begin{algorithmic}[1]
\State \textbf{input}: pool $\{\mathcal{D}_{\tau}\}_{\tau \in \mathcal{T}_{a}}$ of available, unlabeled, meta-training tasks
\State initialize $i=1$
    \While{$i\leq B$}
        \State select meta-training task $\tau$ using  \eqref{bamld_gen}
        \State query oracle to obtain labels $\mathcal{Y}_\tau$
        \State $\mathcal{D}_{\mathcal{T}_s} \leftarrow \mathcal{D}_{\mathcal{T}_s} \cup \mathcal{D}_{\tau}$
        \State $\mathcal{T}_s \leftarrow \mathcal{T}_s \cup  \{\tau\}$
        \State update variational posterior $q(\theta|\mathcal{D}_{\mathcal{T}_s})$ (using VI or Monte Carlo methods)
        \State $i \leftarrow i+1$
    \EndWhile
\end{algorithmic}
\end{algorithm}

\subsection{BAMLD for Gaussian Process Regression and BO}
Following similar steps as \cite{houlsby2011bayesian}, we can directly instantiate BAMLD for Gaussian Process Regression (GPR) and BO. The use of GPs has the computational advantage that the marginalization over the model parameter $\phi_{\tau}$ required to compute the predictive distribution $p(\tilde{\mathcal{Y}}_\tau \,|\, \tilde{\mathcal{X}}_\tau,  \theta)$, which is in turn needed to evaluate the meta-acquisition function (\ref{MI_bamld_gen}), can be obtained in closed form.

To elaborate, the GP model assumes the marginal \cite{rasmussen2003gaussian}    \begin{align}\label{marginalGP}
    p(\mathcal{Y}_\tau \, | \, \mathcal{X}_\tau,\theta) &= \mathcal{N}(\mu_{\theta}(\mathcal{X}_\tau),\tilde{K}_{\theta}(\mathcal{X}_\tau)),
\end{align}
where $\mu_{\theta}(\mathcal{X}_\tau) = [\mu_{\theta}(x_\tau^1),...,\mu_{\theta}(x_\tau^{N_\tau})]$ represents the $N_\tau \times 1$ mean vector, and $\tilde{K}_{\theta}(\mathcal{X})=K_{\theta}(\mathcal{X})+\sigma^2 I$ is the $N_\tau \times N_\tau$ covariance matrix with $[K_{\theta}(\mathcal{X}_\tau)]_{i,j} = k_{\theta}(x_\tau^i,x_\tau^j)$. The model  is parametrized by the shared hyperparameter ${\theta}$ through the mean and kernel functions defined as
\begin{align}\label{meanNN}
    \mu_{\theta}(x_\tau^n) = \Phi^{\mu}_{\theta}(x_\tau^n),
\end{align}
and
\begin{align}\label{covarNN}
    k_{\theta}(x_\tau^n,x_\tau^j) = \frac{1}{2} \exp (-||\Phi^k_{\theta}(x_\tau^n) - \Phi^k_{\theta}(x_\tau^j)||_2^2),
\end{align}
where $\Phi^\mu_{\theta}(\cdot)$ and $\Phi^k_{\theta}(\cdot)$ are neural networks. Using the predictive distribution $p(\mathcal{Y}_\tau \, | \, \mathcal{X}_\tau,\theta)$ in (\ref{marginalGP}) and given a variational distribution $q(\theta|\mathcal{D}_{{\mathcal{T}}_s})$, the meta-acquisition function (\ref{MI_bamld_gen}) can be evaluated as follows. 

Using (\ref{marginalGP}), the second term in (\ref{MI_bamld_gen}) can be approximated as
\begin{align}\label{expect_second}
    &\E_{q({\theta} \,|\, \mathcal{D}_{\mathcal{T}_s})} [\mathrm{H} (\tilde{\mathcal{Y}}_\tau \,|\, \tilde{\mathcal{X}}_\tau, {\theta})] \nonumber\\
    &= \E_{q({\theta} \,|\, \mathcal{D}_{\mathcal{T}_s})}\left[ \frac{1}{2} \log \det(2\pi\mathrm{e}\tilde{K}_{\theta}(\tilde{\mathcal{X}}_\tau)) \right],
\end{align}
where $\det(\cdot)$ denotes the determinant, and the outer expectations can be estimated using samples from the variational posterior as discussed in the previous section. A similar discussion applies to the first term in  (\ref{MI_bamld_gen}). We provide further details in Sec. \ref{sec:Approx}.

\section{Experiments}
In this section, we empirically evaluate the performance of BAMLD for BO.

\subsection{Experimental Setup}
We focus on the problem of optimizing an unknown black box function $g(\cdot)$ as per the maximization $x^{*} = \text{arg max}_{x \in \mathcal{X}} \textrm{g}(x)$, where $x$ is a scalar real quantity. This may correspond, for instance, to the problem of optimizing the location of an aerial base station along a segment of a highway with the goal of maximizing some desired, hard-to-evaluate, performance indicator $g(\cdot)$, such as quality of service. We are interested in obtaining good solutions by using a minimal number of queries on the input $x$. To do so, BO chooses consecutive inputs $x^n$ at which to query function $\textrm{g}(\cdot)$, and observes noisy feedback $y^n$ as $y^n = \textrm{g}(x^n) + \epsilon$, with $\epsilon \sim \mathcal{N}(0, \sigma^2)$.
We choose the class of objective functions \cite{rothfuss2021meta}
\begin{align}\label{g_bo}
    &\textrm{g}_{\alpha_1,\alpha_2,\alpha_3} (x) \nonumber\\
    &= 2 \, w_1 \, p_1(x) + 1.5 \, w_2 \, p_2(x) + 1.8 \, w_3 \, p_3(x) + 1,
\end{align}
where the parameters $\tau = (w_1, w_2, w_3, \alpha_1, \alpha_2, \alpha_3)$ define the task, and we have the unnormalized Cauchy and Gaussian probability density functions
\begin{align}
    p_1(x) &= \frac{1}{\pi(1+||x-\alpha_1||^2)}, \nonumber\\
    p_2(x) &= \frac{1}{2\pi}e^{-\frac{||x-\alpha_2||^2}{8}}, \nonumber\\
    p_3(x) &= \frac{1}{\pi(1+\frac{||x-\alpha_3||^2}{4})},
\end{align}respectively.
The population distribution used to generate meta-training data for a given task $\tau$ is such that $x \sim \mathcal{U}(-10, 10)$ follows a uniform distribution, and $y$ is given by \eqref{g_bo}. Furthermore, the task distribution is such that  the parameters $w_1, w_2, w_3$ are independent and distributed as $w_i \sim \mathcal{U}(0.6, 1.4)$ for $i=1,2,3$, and the parameters $\alpha_1, \alpha_2, \alpha_3$ are distributed as $\alpha_1 \sim \mathcal{N}(-2,0.3^2)$, $\alpha_2 \sim \mathcal{N}(3,0.3^2)$, and $\alpha_3 \sim \mathcal{N}(-8,0.3^2).$

During meta-testing, the meta-trained GP is used as a surrogate function to form a posterior belief over the function values, and, at each iteration of BO, the next query point is chosen to maximize the upper confidence bound (UCB) acquisition function, which is designed to balance exploration and exploitation \cite{garivier2011upper}. The mean and kernel parameters of the surrogate are then updated, and the procedure is repeated. This setting conforms to \cite{rothfuss2021meta}, which, however, considers only standard passive meta-learning. 

\subsection{Posterior Approximation}\label{sec:Approx}
In our implementation, we have estimated the expectations with respect to the variational posterior using SVGD as in \cite{rothfuss2021pacoh}. SVGD  amounts to maintaining $P$ samples $\{\theta_p\}_{p=1}^{P}$, using which we can obtain the estimates
\begin{align}
    &\E_{q({\theta} \,|\, \mathcal{D}_{\mathcal{T}_s})} \mathrm{H} (\tilde{\mathcal{Y}}_\tau \,|\, \tilde{\mathcal{X}}_\tau, {\theta}) \nonumber\\
    &\approx \frac{1}{2} \frac{1}{P} \sum_{p=1}^P \left[ \frac{1}{2} \log \det (2\pi\mathrm{e}\tilde{K}_{\theta_{p}}(\tilde{\mathcal{X}}_\tau)) \right],
\end{align}
and 
\begin{align}\label{expect_second}
    &\mathrm{H} (\tilde{\mathcal{Y}}_\tau \,|\, \tilde{\mathcal{X}}_\tau, \mathcal{D}_{\mathcal{T}_s}) \nonumber\\
    &\approx \E_{\hat{p}(\tilde{\mathcal{Y}}_\tau \,|\, \tilde{\mathcal{X}}_\tau,  \theta)} [-\log \hat{p}(\tilde{\mathcal{Y}}_\tau \,|\, \tilde{\mathcal{X}}_\tau,  \theta)],
\end{align}
in which $\hat{p}(\tilde{\mathcal{Y}}_\tau \,|\, \tilde{\mathcal{X}}_\tau,  \theta) = (1/P) \sum_{p=1}^P p(\tilde{\mathcal{Y}}_\tau \,|\, \tilde{\mathcal{X}}_\tau, {\theta}_p)$ is a mixture of multivariate normal distributions. 

\subsection{Architectures and Hyperparameters}
The neural networks $\Phi^\mu_{\theta}(x_\tau^n)$ and $\Phi^k_{\theta}(x_\tau^n)$ are comprised of 2 hidden layers each, with $(32,32)$ hidden neurons. We use tanh as a non-linearity. The optimization is performed using Adam over $10000$ training iterations. The learning rate is set to $0.001$ and is decreased every $1000$ epochs. We use an RBF kernel for SVGD. The initial task pool size is $|\mathcal{D}_{\tau_a}| = 20$, each with	   $|\mathcal{X}_\tau| = 40$  samples. The number of acquired tasks  is $B = 12$ and we use $40$ samples for BO.

All experiments have been conducted on CPU-only machines on Google Colab. We have approximately used $150$ hours of runtime. All results are averaged over $5$ independent trials.

\subsection{Results}

\begin{figure}[tbp]
\centering
\includegraphics[width=0.9\linewidth]{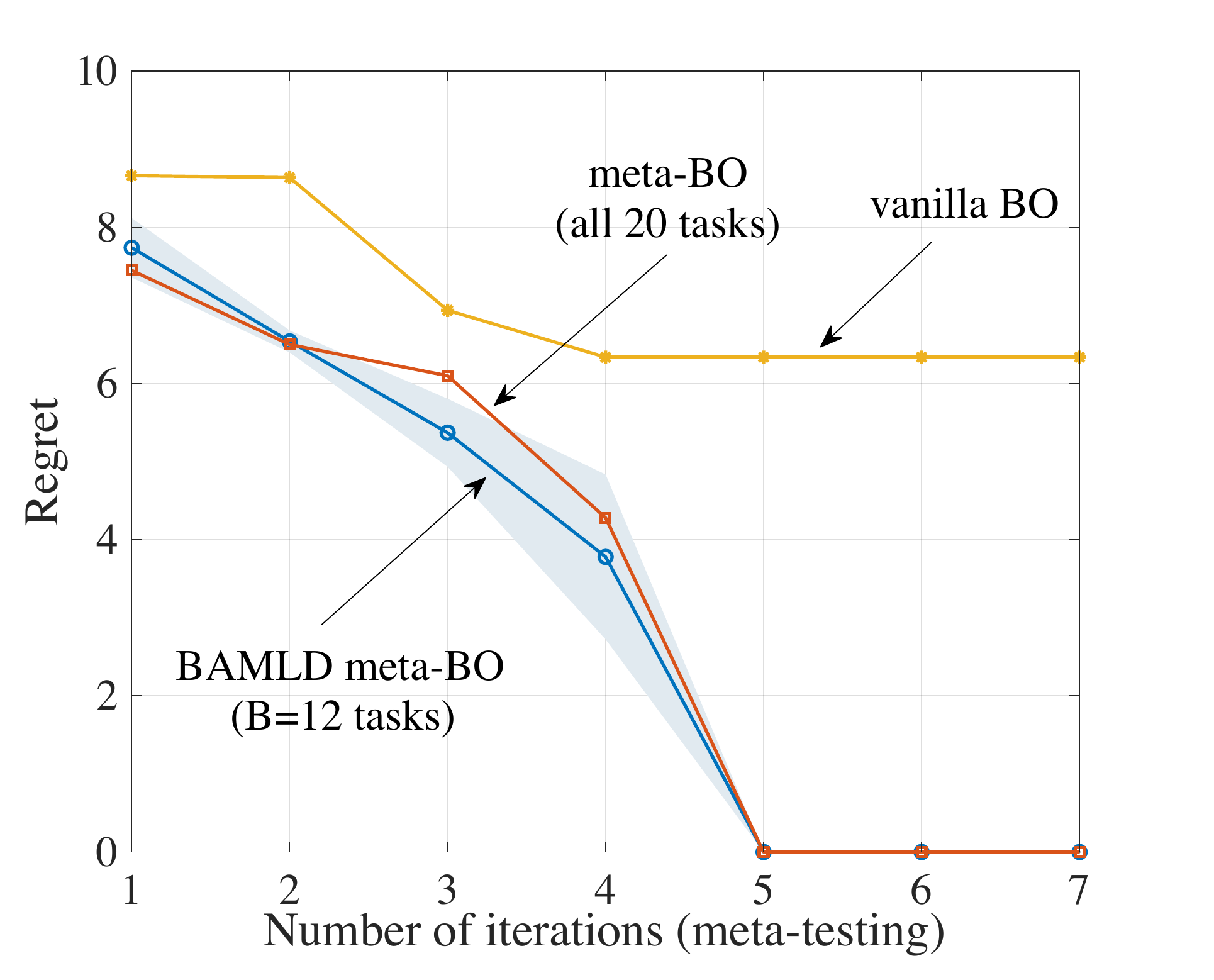}
\caption{Regret for BO during meta-testing as a function of the number of iterations. For BAMLD BO, the initial size of the pool of available task is set to $|\mathcal{D}_{\tau_a}| = 20$, from which $B=12$ tasks are selected. The number of samples is set to $|\mathcal{X}_\tau| = 40$. The results are averaged over $5$ independent seeds.}
\vspace*{-5mm}
\label{bo_one}
\end{figure}

We report the regret $\mathcal{R}$, which is defined as
\begin{align}
    \mathcal{R} = \underset{x \in \mathcal{X}}{\text{max}} \,\, \textrm{g}(x) - \underset{t = 1,2,...}{\text{max}} \,\, \textrm{g}(x^{t^*}),
\end{align}
where $x^{t^*}$ denotes the best solution obtained in the optimization procedure so far, i.e., $g(t^*)=\min_{t}g(x^{t})$. As benchmarks, we consider the \textit{vanilla BO} scheme that neglects meta-training data and assumes a GP prior with mean and a squared-exponential kernel with parameters as in \cite{srinivas2009gaussian}; as well as an ideal \textit{meta-BO} scheme for which the mean and kernel are meta-optimized in the offline phase using all $|\mathcal{T}_{a}| = 20$ meta-training data sets \cite{rothfuss2021meta}. Meta-testing is done for the proposed BAMLD meta-BO scheme after selecting $B = 12$ meta-training tasks out of the pool of $20$ tasks. 

Vanilla BO is seen to have the poorest performance among all schemes, getting stuck in a sub-optimal solution after a few iterations.
Conversely, meta-BO finds good solutions quickly, requiring only $5$ function evaluations before the regret is minimized. 
A comparable regret is achieved with BAMLD meta-BO. However, BAMLD meta-BO uses only  $12$ out of $20$ tasks, as opposed to all tasks as the ideal meta-BO scheme, indicating that it is capable of achieving a positive transfer for BO with only a fraction of the available tasks.

\section{Conclusions}

This paper introduced BAMLD, a method for information-theoretic active meta-learning. An instantiation for BO was detailed, and
BAMLD was shown to decrease the number of required meta-training tasks for BO, and still achieve a positive transfer. This result is encouraging for many applications that depend on black-box optimization such as optimization of system resources in wireless networks.


\bibliographystyle{IEEEtran}
\bibliography{litdab.bib}

\end{document}


%

%

\onecolumn
\aistatstitle{Supplementary Materials}

\section{Implementation Details}

All experiments have been conducted on CPU-only machines on Google Colab. We have approximately used $150$ hours of runtime. Code will be made available. All results are averaged over $5$ independent trials.

\begin{table*}[!h]
\centering
\caption{Architecture and training hyper-parameters}
\label{table2}
\begin{tabular}{lll}
\hline
Hyper-parameters & Figs. 2,3,4 & Fig. 5     \\ \hline
Number of hidden layers, $\Phi^{\mu}_{\theta}(\cdot)$ & $2$ & $2$   \\ \hline
Number of hidden layers, $\Phi^{k}_{\theta}(\cdot)$ & $2$ & $2$   \\ \hline
Number of hidden neurons, $\Phi^{\mu}_{\theta}(\cdot)$ & $(32,32)$ & $(32,32)$   \\ \hline
Number of hidden neurons, $\Phi^{k}_{\theta}(\cdot)$ & $(32,32)$ & $(32,32)$   \\ \hline
Non-linearity & tanh & tanh  \\ \hline
Task mini-batch size     & $2$ & $2$  \\ \hline
Number of meta-training iterations     & $10000$  & $10000$  \\ \hline
Learning rate     & $0.001$  & $0.001$  \\ \hline
Learning rate decay (multiplier applied after every 1000 steps)     & $1$  & $1$  \\ \hline
Optimizer     & Adam  & Adam   \\ \hline
SVGD kernel     & RBF  & RBF   \\ \hline
Mean and kernel update iterations for BO (after each query)   & NA  & $100$   \\ \hline
\end{tabular}
\end{table*}

\begin{table*}[!h]
\centering
\caption{Meta-learning hyper-parameters}
\label{table3}
\begin{tabular}{llll}
\hline
Hyper-parameters & Fig. 2,3 & Fig. 4 & Fig. 5          \\ \hline
Initial pool size      & $|\mathcal{D}_{\tau_a}| = 20$ & $|\mathcal{D}_{\tau_a}| = 20$   & $|\mathcal{D}_{\tau_a}| = 20$     \\ \hline
Number of samples	       & $|\mathcal{X}_\tau| = 40$ & $|\mathcal{X}_\tau| = 40$   & $|\mathcal{X}_\tau| = 40$  \\ \hline
Number of acquired tasks   & NA   & $B = 12$ & $B = 12$  \\ \hline
Number of samples for BO   & NA   & NA  & $40$  \\ \hline
\end{tabular}
\end{table*}

\vfill